\documentclass{IEEEtran}
\usepackage{verbatim}  
\usepackage{algorithm}
\usepackage{algpseudocode}
\usepackage{eurosym}
\usepackage{float}
\usepackage{amssymb}
\usepackage{amsmath}
\usepackage{graphicx}
\usepackage[T1]{fontenc}
\usepackage{flushend}
\usepackage{chngcntr}
\usepackage{url}
\usepackage{xspace}
\usepackage{listings}
\usepackage{fancyhdr}
\usepackage[normalem]{ulem}
\usepackage{subfig}
\usepackage{enumitem}
\usepackage{tikz}

\counterwithin{paragraph}{subsection}

\usepackage{graphicx}

\begin{document}

\title{Risk Assessment, Prediction, and Avoidance of Collision in Autonomous Drones}

\author{\
\IEEEauthorblockN{Anamta Khan}

\IEEEauthorblockA{
	CISUC, Department of Informatics Engineering, University of Coimbra, Portugal
	\\
		anamta@dei.uc.pt 
}
}
    
\maketitle

\begin{abstract}

Unmanned Aerial Vehicles (UAVs), in particular Drones, have gained significant importance in diverse sectors, mainly military uses. Recently, we can see a growth in acceptance of autonomous UAVs in civilian spaces as well. However, there is still a long way to go before drones are capable enough to be safely used without human surveillance. A lot of subsystems and components are involved in taking care of position estimation, route planning, software/data security, and collision avoidance to have autonomous drones that fly in civilian spaces without being harmful to themselves, other UAVs, environment, or humans. 
The ultimate goal of this research is to advance collision avoidance and mitigation techniques through quantitative safety risk assessment. To this end, it is required to identify the most relevant faults/failures/threats that can happen during a drone's flight/mission. The analysis of historical data is also a relevant instrument to help to characterize the most frequent and relevant issues in UAV systems, which may cause safety hazards. Then we need to estimate their impact quantitatively, by using fault injection techniques.  
Knowing the growing interests in UAVs and their huge potential for future commercial applications, the expected outcome of this work will be helpful to researchers for future related research studies. Furthermore, we envisage the utilization of expected results by companies to develop safer drone applications, and by air traffic controllers for building failure prediction and collision avoidance solutions.

\end{abstract}

\thispagestyle{empty}
\IEEEpeerreviewmaketitle

\section{Introduction}
\label{sec:intro}

Unmanned aerial vehicles (UAVs), and more particularly drones, are aircrafts without a pilot on board, which are part of a more extensive system, the unmanned aircraft system (UAS) \cite{UAS_SESAR}. The UAS consists of the UAV itself, sensors, ground control station and support services. Drones can be controlled by a ground station pilot or being autonomous, having a mission payload that does not require human surveillance \cite{UAVDesc2015}. UAVs are considered to be one of the major technological advancements \cite{UAVDesc2020} because of their wide range of uses and their application in multiple sectors, such as item deliveries, disaster management, rescue operations, geographic mapping, safety inspections, crop monitoring, hurricanes and tornadoes monitoring, law enforcement and border control \cite{applications}. They even have a role in more unique missions such as Visual Inspection of Structures \cite{visual2014}, and seem to have an even higher potential for the coming years. 

The overall drone market is expected to grow to USD $52$ billion by 2025 in comparison to USD $11$ billion in 2016 \cite{dn, insight} and it is clear that the adaptation of autonomous drones in civilian spaces has a significant role to play in this expected growth. This is mainly due to their flying ability, great scope of applications, simplicity in terms of mechanical design, relatively low price, and also their potential to perform tasks that are costly to be performed by humans or that threaten human lives \cite{droneEmergency}.

Although drones have a high potential for a wide range of uses in civilian airspace, and can bring affordable life-changing applications, the enormous complexity of the entire drone system also brings a severe threat of failures that may result in damage to the drones and, more important, may cause environmental damages (e.g., damage in buildings, cars, trees) and ultimately may lead to human casualties, specially with the use of autonomous drones (i.e., UAVs that are able to fulfill a mission without a pilot or human surveillance.).

Quantitative safety risk analysis and assessments \cite{uavsec} provides deep insight and understanding of autonomous drones' mid-flight issues that can cause collisions. This helps in building safer drones and more effective mitigation and collision avoidance strategies and techniques.  
Safety risk assessment in UAVs is usually addressed using analytical approaches in the literature. Prominent examples of such approaches are the methodologies developed by European ECSEL JU projects such as CORUS that have developed SORA (Specific Operations Risk Assessment) \cite{capitan2019risk} and MEDUSA (Methodology for U-Space Safety Assessment) \cite{barrado2020u}. Despite the merit of these approaches, it is quite difficult, if not impossible, to consider all parameters, involved in safety assessment, into consideration, due to the complexity of UAV systems.   

Fault injection is a widely accepted technique for the evaluation of the impact of faults/failures/threats in computer systems and is a natural choice for experimental evaluation of drone systems dependability \cite{FIprocess}. This thesis aims to demonstrate a practical way of performing fault injection in drone systems, through the injection of realistic faults and failures in actual systems running real drone missions in a simulation environment. To support the study throughout, we will look at previous research to better understand the concepts involved, analyze the related work, and make use of well-proven databases, techniques and methods if necessary. Furthermore, we will make available the datasets or tools created throughout this research to be used by other researchers.
\section{Background and Related Studies}
\label{sec:relatedstudies}

This section discusses some previous works similar to the studies that are planned along the lines of this PhD thesis. These include the approaches that are used for UAV risk assessments and techniques for experimentation.

The use of drones has been increasing over time and we are currently experiencing a rapid increase of research effort related to these vehicles. An important focus of the research work around UAVs is related to the assessment of the safety and security aspects, which is considered an essential aspect for the success of drones applications in the near future. There are two main phases of an autonomous drone's mission, i) Strategic phase, where the mission is planned and initial requirements are completed and ii) Tactical phase which is the mission flight phase where monitoring happens. This PhD thesis is focused on the tactical conflicts because they may occur during flight, despite the strategic phase efforts to provide conflict-free missions. Prediction of conflicts combines the planned mission with the current position and motion of the drones, and uses the result to generate a probabilistic trajectory, again considering other factors and the desired separation. Conflict detection is based on the probability of overlap of these probabilistic trajectories exceeding some pre-defined acceptable value (i.e., minimum separation distance).

Being highly accepted by the research industry, risk or safety assessment is a major method of research used to asses risks, which includes the identification of faults and failures that may occur. Thus, a large number of researches based on this type of assessment can be found and consequently, we can also find a great amount of works on risk assessment for drones based on different scenarios and components. Although there are certain risk assessment models that are more widely used, there is a lack of an accepted model throughout the research community in this area. This is mainly due to the difficulty in providing a general model, given the amount of different scenarios a drone can go through, the risks that can originate from them and the scalability these scenarios can present, depending on the number of drones. Two different methods for risk assessment are presented in \cite{guglieri2014operational} that are mainly based on the damages caused by possible drones' failures.

In a recent study \cite{Vulnerability2013}, the authors suggest an approach to risk assessment for UAVs based on a vulnerability study of cyber attacks on UAVs. This paper \cite{Vulnerability2013} provides numerous risks of two specific UAVs (i.e., MQ-9 Reaper and AR Drone) using known cyber attacks (such as GPS spoofing) and an analysis of the communication infrastructures that are based on the drones' exposure, communication systems, storage media, sensor systems and fault handling mechanisms. With this approach, the risk of cyber attacks can be calculated for each individual drone using past data. A similar approach considering four UAVs is discussed in this paper \cite{reliability2017}, where past failure rate history data from these four UAVs was used to assess the reliability of the drones.

Similarly, the use of both qualitative and quantitative risk analysis methodologies for cloud based drones safety over the same network/system was proposed in \cite{allouch2019qualitative}. The qualitative analysis is based on a combination of ISO 12100 (i.e., focused on risk analysis and assessment of any machine) and ISO 13849, which is an internationally registered standard for performance level of machines and concerns all levels of manufacturing. The second analysis in \cite{allouch2019qualitative} (quantitative) is done by applying a Bayesian network-based technique over data collected from previous studies with the objective of analyzing the relationships between the risk of drone crashes and their causes.

The intentional injection of faults and failures in computer systems is a great way to evaluate the effect of faults and failures in such systems, as well as evaluating fault handling mechanisms and study error propagation \cite{faultInjection2007}. In a recent study \cite{designFI2019}, the authors proposed an HLA architecture and injection tool based on simulation model for analyzing fault injections in UAVs. This tool allow performing experimentation on some types of failures such as freezing GPS values, allowing the comparison of the results from the same mission with no fault or failure injected. This is achieved by comparing the logs or real-time data. A similar approach is used in \cite{chandhrasekaran2010fault}, where a fault injection tool is used to inject faults associated with yaw, pitch, roll and wind speed parameters in a Hardware in the Loop (HITL) simulation environment. Similarly in \cite{mendes2018effects}, the authors analyze the effects of GPS spoofing on drones through a series of tests in a Hardware in the Loop (SITL) simulated environment.

A UAVs safety assessment approach is presented in \cite{deligne2012ardrone}, where a series of attacks, namely, DoS, Wireless Injection, and Video Capture Hijack are used on a real commercial drone to show how easily one can remotely control this type of vehicle. In a similar study \cite{gordon2019security}, a De-Authentication attack is emulated to demonstrate that anyone with access to a computer could potentially take down a drone. Another similar study is reported in \cite{yihunie2020assessing}, the paper aims to assess the security of commercially available drones. The authors present four security vulnerabilities found in two drones and exploit them through a series of attacks - De-Authentication attack, File Transfer Protocol Service Attack, Radio Frequency Replay Attack and a Custom Made Controller Attack. 
\section{Research Questions}
\label{sec:researchquestions}

\begin{figure}[t]
  \centering
  \includegraphics[width=\linewidth]{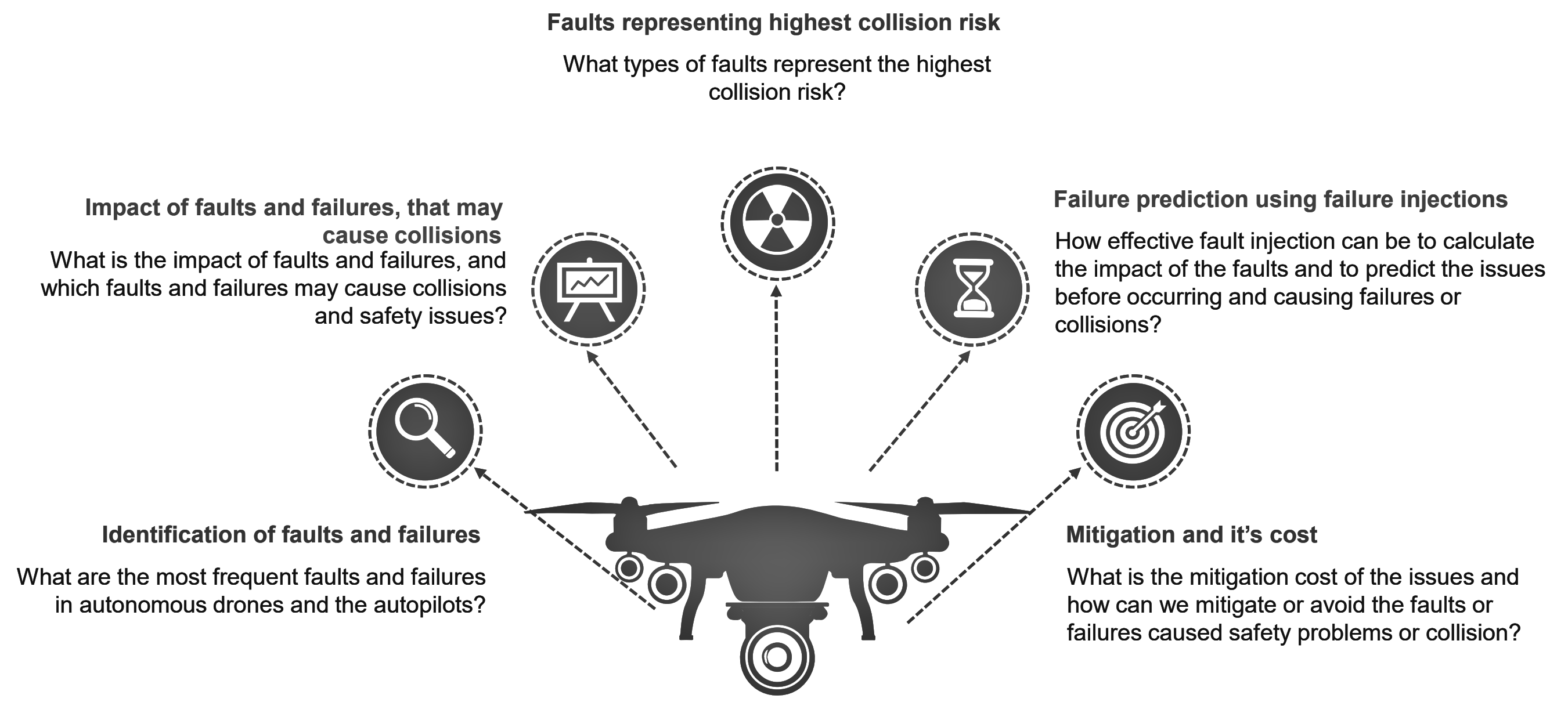}
  \caption{Research Focus.}
  \label{fig:focus}
\end{figure}

The research questions of this PhD thesis are in line to make drones more safe and avoiding possible collisions among them or with other objects. As can be seen in Figure \ref{fig:focus}, the focus of this study and research questions are defined in a way that leads, step by step, from the identification to possible prediction and avoidance of collisions.
\begin{enumerate}
\item What are the most frequent faults and failures in UAVs, autonomous drones and the autopilots?
\item What is the impact of each fault and failure? Which faults and failures may cause collisions and safety issues?
\item What types of faults or failures represent the highest collision risk?
\item How effective fault injection technique can be to calculate the impact of the UAVs' faults/failures?  Can the obtained results be used to to predict the issues before occurring and causing collisions? If so, how?
\item What is the mitigation cost of the failures and how can we apply mitigation techniques to avoid (failures caused) safety problems or collision?
\end{enumerate}

\section{Methodology}
\label{sec:methodology}

\begin{figure}[h]
  \centering
  \includegraphics[width=\linewidth]{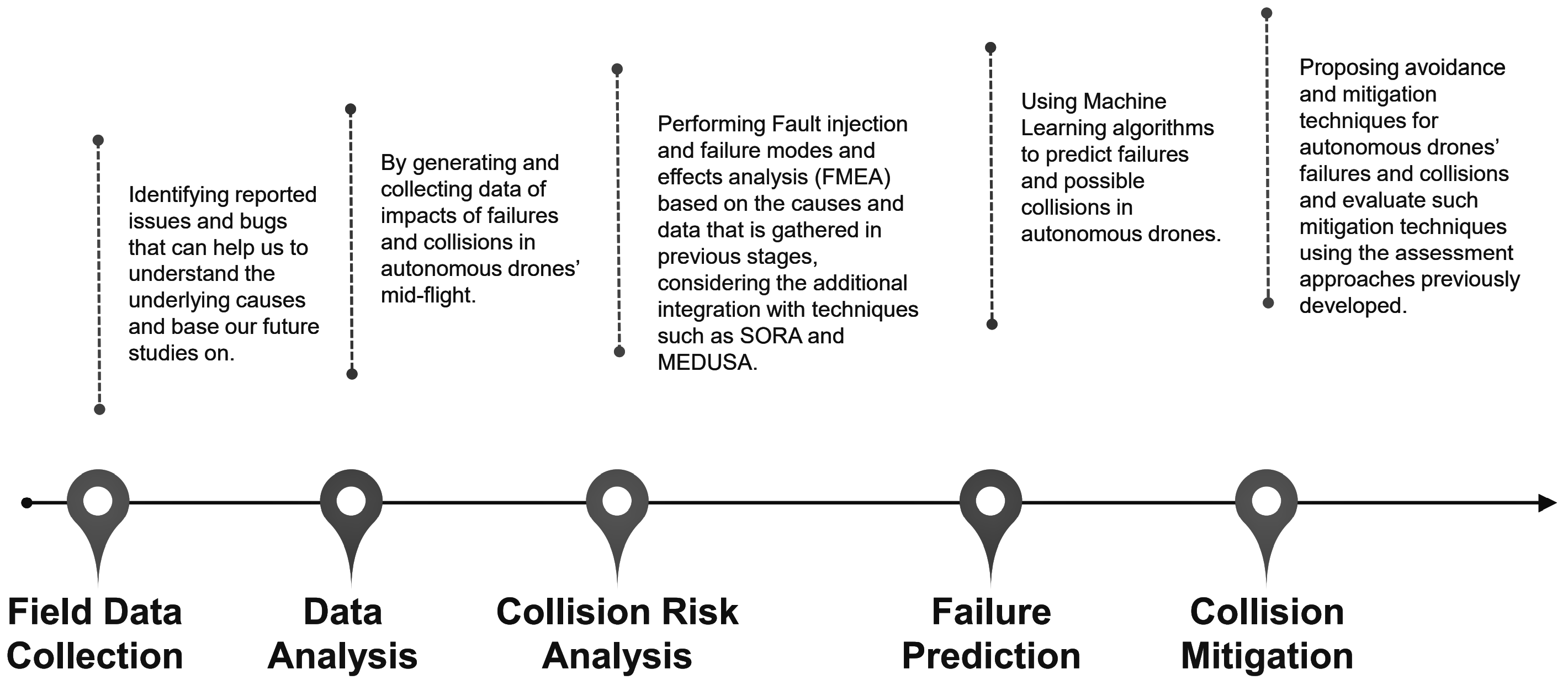}
  \caption{Research Process.}
  \label{fig:objectives}
\end{figure}

The objectives of this PhD thesis is to help in making drones specifically autonomous drones safe to be able to be used in city skies, which can help to achieve many everyday tasks autonomously. The proposed methodology is largely based on experimental approaches and on the integration of experimental results with analytical methods currently used to asses risk and safety in UAVs. As can be seen in Figure \ref{fig:objectives}, the envisioned approach and objectives includes the following steps:

\begin{enumerate}
\item Identifying causes from real scenarios: Identifying reported issues and bugs that can help us to understand the underlying causes and base our future studies on.
\item Data generation / collection: By generating and collecting data of impacts of failures and collisions in autonomous drones' mid-flight.
\item FMEA: Performing failure modes and effects analysis (FMEA) based on the causes and data that is gathered in previous stages, considering the additional integration with techniques such as SORA and MEDUSA.
\item Prediction: Using Machine Learning algorithms to predict failures and possible collisions in autonomous drones.
\item Avoidance / Mitigation: Proposing avoidance and mitigation techniques for autonomous drones' failures and collisions and evaluate such mitigation techniques using the assessment approaches previously developed.
\end{enumerate}

These steps are mainly focused for air-traffic control systems, thus the studies in this PhD thesis will largely target individual autonomous drones and the ones over a same network to avoid collisions and failures.

\subsection{Experimental Environment}
\label{sub:setup}

\begin{figure}[t]
    \centering
    \includegraphics[width=\linewidth]{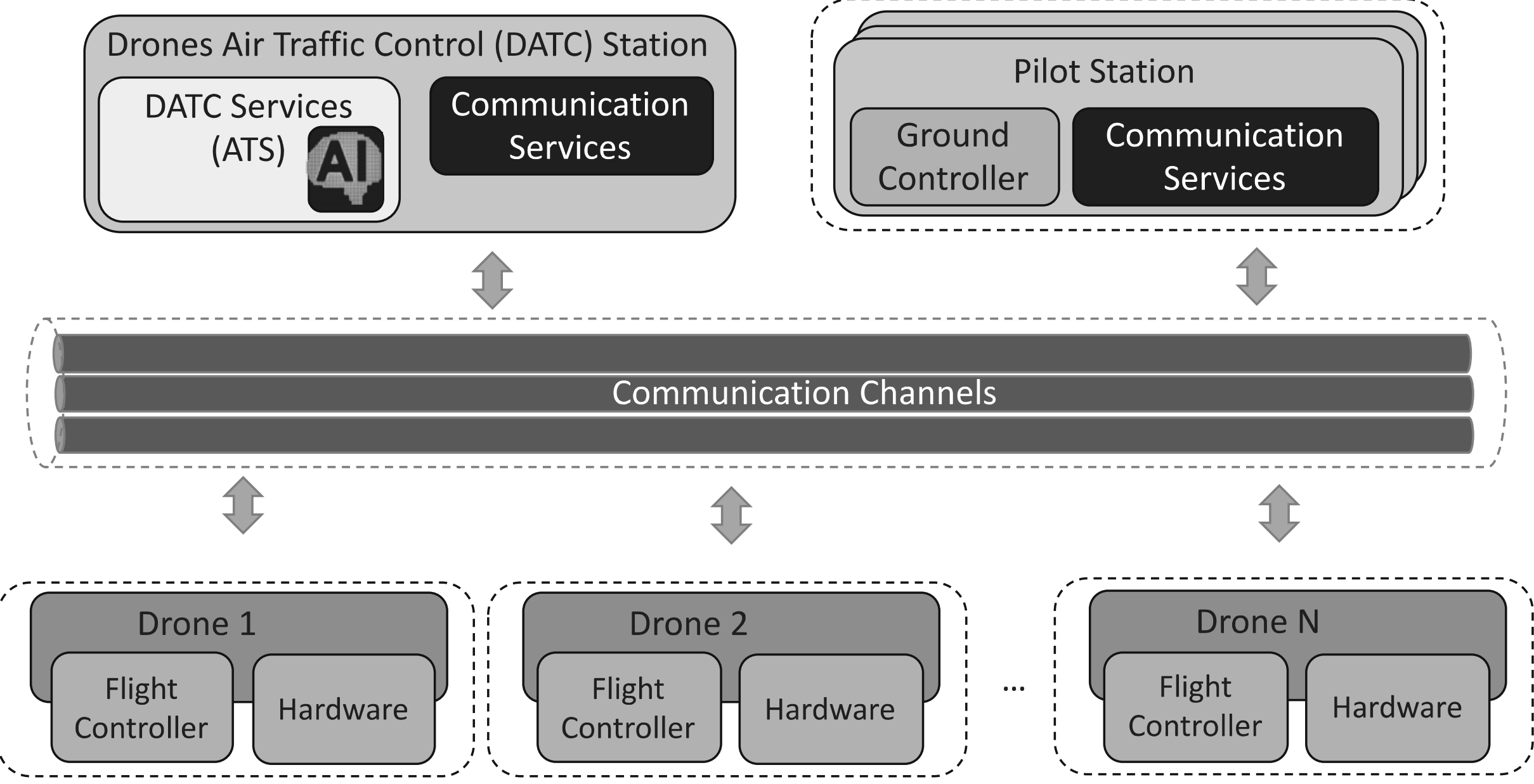}
    \caption{High level view of the Experimental Setup.}
    \label{fig:droneInfrastructure}
\end{figure}

Autonomous drones in a crowded sky (i.e., within smart cities) cannot fully operate by themselves in a safe manner. They are a component within a more extensive system, ideally, including a unmanned aircraft system (UAS) \cite{uas} and a drone air traffic control (DATC) station(s) \cite{bubbles}. Figure \ref{fig:droneInfrastructure}, presents a general view of this extensive system. A UAS is composed of the following principal components \cite{sharma2020communication}:

\begin{itemize}
    \item A drone, composed by both hardware - namely a body, a power supply, a flight controller board, payloads (e.g., sensors, GPS, etc. that may vary depending on the missions to be accomplished), and the necessary actuators - and software - mainly the flight controller which can be composed by a firmware, a middleware and an operating system;
    \item A pilot station with a ground controller from which the pilot interacts with the UAV to control and monitor it;
    \item A communication equipment required to establish communication between the drone and the pilot station; 
\end{itemize}

In this context, a drone air traffic control (DATC) station offer services to minimize the collision by determining the necessary minimum separation distance and the criteria and methodologies to maintain that separation distance between drones. These services are used to avoid both strategic conflicts and tactical conflicts. Strategic conflicts occur before take-off and strategic level services are intended to result in conflict-free missions. Tactical conflicts may occur during flight, in spite of the strategic phase effort to provide conflict-free missions. Prediction of conflicts combines the planned mission with the current position and motion of the drones and uses the result to generate a probabilistic trajectory, again considering other factors and the desired separation. Conflict detection is based on the probability or overlap of these probabilistic trajectories exceeding some pre-defined acceptable value (i.e., minimum separation distance).

This study aims to simulate this context with the use of open source autopilot PX4 using Gazebo and Pxhawk, in combination with DroneKit, a python framework that allows direct communication with MAVLink vehicles. We use this framework to establish default missions that the simulated UAVs can run and. The proposed experimental environment will provide an effective way to observe the impact of faults through the output logs of the executed simulations.

\subsection{Challenges}
\label{sub:challenges}

The study posses some expected challenges as well, with usage of drones being relatively new in public scenarios, there is not much relevant data available, although they have been used by army from a long time but their private and not easily accessible. But to overcome that the study will use simulated environments with faults and failure injections to emulate the required data.

Another prospective challenge that the study posses is validity of this simulated data, because most of the data generated and collected is based on the logs of simulations, to that end, the study will identify some scenarios to verify them on real drone to validate if the simulated data is co-relates with the real world data.
\section{Expected Results}
\label{sec:expectedresults}

Following the objectives, the end results of this PhD are expected to be produce results that will be specifically beneficial for unmanned aircraft system (UAS) and drone air-traffic control (DATC) systems. This PhD thesis aims to produce multiple studies, starting with analyzing the reported issues for autopilots will help in targeting the most critical issues that needs to be addressed to increase safety of the drones, environment and to the humans, for them to be used in the civilian spaces.

Followed by identification and assessments of faults and failures in UAVs, this study proposes predictions and avoidance techniques backed by i) Existing data ii) Data generated by the existing tools or the tools created during this study and iii) The experimentation and tests performed on autonomous drones and autopilots using simulations and real drones as examples. 

\bibliographystyle{IEEEtran}
\bibliography{references.bib}

\end{document}